\documentclass[aps,pra,preprint,groupedaddress,showpacs]{revtex4-1}
\usepackage{amssymb}
\usepackage{amsmath}
\usepackage{dcolumn}
\usepackage{bm}
\usepackage{graphicx}
\usepackage{mathrsfs}

\begin{document}

\title{Imaging around corners with single-pixel detector by computational ghost imaging}
\author{Bin Bai}
\author{Jianbin Liu}
\author{Yu Zhou}
\email{zhou1@mail.xjtu.edu.cn}
\author{Songlin Zhang}
\author{Yuchen He}
\author{Zhuo Xu}

\begin{abstract}
We have designed a single-pixel camera with imaging around corners based on computational ghost imaging. It can obtain the image of an object when the camera cannot look at the object directly. Our imaging system explores the fact that a bucket detector in a ghost imaging setup has no spatial resolution capability. A series of experiments have been designed to confirm our predictions. This camera has potential applications for imaging around corner or other similar environments where the object cannot be observed directly.

\end{abstract}

\maketitle

\address{Electronic Materials Research Laboratory, Key Laboratory of the Ministry of Education \& International Center for Dielectric Research, Xi'an Jiaotong University, Xi'an 710049, China}
\address{Key Laboratory of Quantum Information and Quantum Optoelectronic Devices, Xi'an Jiaotong University, Xi'an 710049, China}
\section{introduction}

Ghost imaging (GI) was first realized by Pittman \emph{et.al.} in 1995 by employing entangled photon pairs generated from spontaneous parametric down conversion \cite{1}. At the beginning of the 21st century, it was realized that GI could also be performed with classical light \cite{2,3,4,5,thermal light,7,8,9,pseudo-thermal light}. In this system, a thermal light beam is split into two beams by a beam splitter. One beam is projected onto an object. The light reflected from or passing through the object is collected by a bucket detector with no spatial resolution. The other beam is sent to a detector with spatial resolution. Finally the image is reconstructed by correlating the signals from these two detectors. The possible applications of GI in military and civilian have been studied by many groups, such as high-order correlation ghost imaging \cite{12,hgi}, lensless ghost imaging \cite{ott,18,19,20,21,22}, ghosting imaging through turbulence \cite{lli,hgii,23,38}. Nevertheless, a charged coupled device (CCD) is needed to measure the intensity fluctuations in a typical thermal light GI, because the intensity fluctuations with spatial resolution of a typical thermal light can not be predicted. Meanwhile, a beam splitter is also needed. These factors limited the development of the practical application of GI system in daily life.

Computational ghost imaging (CGI) was proposed in 2008 by Shapiro \cite{cgi}. The intensity fluctuations of light in CGI are artificially modulated. The random, yet known, speckle patterns are projected onto an unknown object. The spatial information of detector signals is not needed, therefore a photodiode (PD) can be used to replace a CCD \cite{25}. The need for the beam splitter and camera is removed in CGI, which helps GI system become simpler and more applicable. The practical applications of CGI are developed by more and more groups \cite{cgii,28,29,30,zeng,sp,rtcgi,33,34,3dcgi}. The speed and quality of imaging have both been improved, such as the real-time video with the single-pixel detectors \cite{rtcgi} and the improvement of the signal-to-noise ratio for different systems \cite{33,34}. Moreover, some features of CGI which the traditional camera does not have have been also proposed and realized, such as 3D imaging of CGI \cite{3dcgi}, imaging of CGI through turbulence and against scattering \cite{28,37}. They make CGI more significant in real life.

In this paper, we studied a feature of CGI that imaging an object without looking at it directly, which can be thought as a single-pixel camera with imaging around corners. This camera is based on the technique of single-pixel detector with CGI system. Traditional camera cannot obtain the image when it dose not see the object directly. However, the single-pixel camera with imaging around corners can get the image under the same condition. Other group also has achieved imaging with similar ability \cite{auto}. Their experiment is based on the speckle autocorrelation and  phase-retrieval algorithm. The scattering media is needed as lens. We use a different approach to achieve the same idea and only a single-pixel detector is needed.

The article is organized as follows. In Sec.$\mathrm{\uppercase\expandafter{\romannumeral2}}$, we report the experimental results of CGI and imaging around corners with the single-pixel detector, respectively. The discussions about the physics of the single-pixel camera with imaging around corners are in Sec.$\mathrm{\uppercase\expandafter{\romannumeral3}}$. Section $\mathrm{\uppercase\expandafter{\romannumeral4}}$ summaries our conclusions.

\section{Experiment}

\begin{figure}[htb]
    \centering
    \includegraphics[width=70mm]{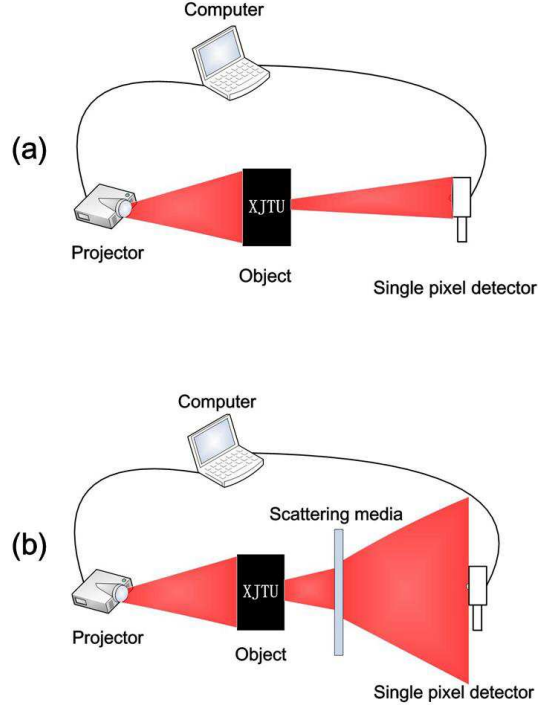}
    \caption{Experimental setup of CGI. (a) is the basic scheme of CGI. A projector is controlled by computer which projects a series of speckle patterns onto an object. The object is transparent letters ``XJTU''. The transmitted light is received by a single-pixel detector. (b) is the  scheme of CGI with scattering media. A ground glass is placed between the object and the detector. The light transmitted through the object will be disturbed by this media.}
    \label{CGI}
\end{figure}

\begin{figure}[htb]
    \centering
    \includegraphics[width=70 mm]{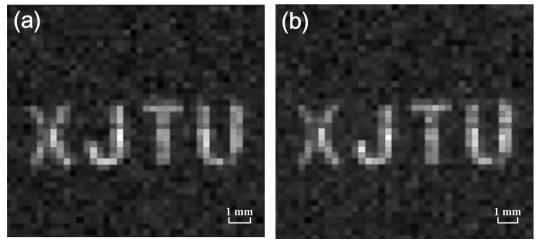}
    \caption{The image of the letters ``XJTU'' in CGI. (a) is the result of letters ``XJTU'' without scattering media. (b) is the image in CGI with scattering media. Two results are both calculated by 18,000 random patterns.}
    \label{XJTU}
\end{figure}

In this article, a series of experiments are designed to study the single-pixel camera with imaging around corners. The first experiment is shown in Fig. \ref{CGI} (a), which is the basic scheme of CGI with a single-pixel detector. The light source is a projector (BenQ MW526). Speckle patterns, controlled by computer, are projected onto an object. The object is a transparent pattern ``XJTU'' and the size of each letter is 0.2 cm$\ast$0.3 cm. The distance between the projector and object is 30 cm. The transmitted light through the object is received by a single-pixel detector (Thorlabs DET36A/M). The function of the detector is to obtain the intensity of light in one position. Figure \ref{XJTU}(a) shows the image of the object in the common CGI.

To test CGI against scattering, a ground glass is placed between the object and the detector as shown in Fig. \ref{CGI} (b). Light through the scattering media will be scattered in all directions. Figure \ref{XJTU}(b) shows the image of XJTU under this condition. The signal-to-noise ratio (SNR) of Fig. \ref{XJTU}(a) and \ref{XJTU}(b) are 18.02 \emph{dB} and 17.92 \emph{dB}, respectively. The results indicate that the quality of the image is nearly the same as that of the image in CGI without scattering media. SNR is defined as $SNR=\bar{s}^{(2)}/\sigma _{n}$, where $ \bar{s}$ is average of the signal intensity, and $ \sigma _{n} $ is the variance of the background intensity \cite{29}. The difference between the average intensity of the bright and dark regions of images is regarded as the signal, and the variation of dark background is considered as the noise.

\begin{figure}[htb]
    \centering
    \includegraphics[width=70mm]{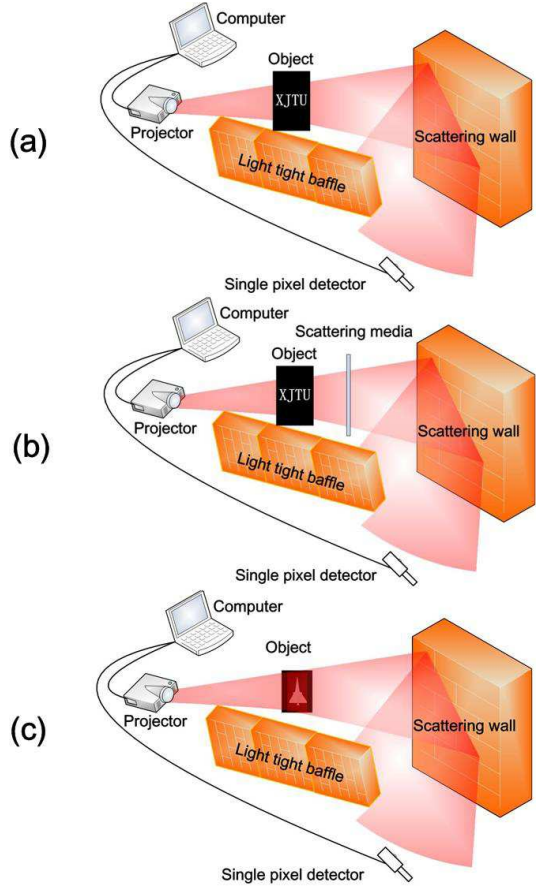}
    \caption{The scheme of single-pixel camera with imaging around corners. (a) is the basic scheme. The transparent letters ``XJTU'' are imaged. A light tight baffle between the object and the single-pixel detector is used for preventing the detector obtaining the light from the object directly.  (b) is the scheme with scattering media. (c) is the scheme of imaging a three-dimensional object with rough surface. This diffuse object is a common toy for children (a plane model).}
    \label{NLOS1}
\end{figure}

\begin{figure}[htb]
    \centering
    \includegraphics[width=85 mm]{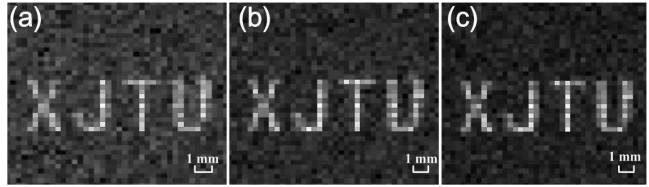}
    \caption{The image of the letters ``XJTU'' by the single-pixel camera with imaging around corners. (a) is the result of letters ``XJTU'' without scattering media. (b) is the image through a rotating ground glass. (c) shows the result when the ground glass is replaced by the white paper. Two pieces of paper are placed at two different locations. The thickness of paper is 0.05 mm.}
    \label{NLOSXJTU}
\end{figure}

Based on the above results, this imaging technology is exploited as a single-pixel camera with imaging around corners. The characteristic of CGI against scattering is employed. Figure \ref{NLOS1}(a) shows the experimental scheme of the single-pixel camera with imaging around corners. The same object ``XJTU'' is used. Speckle patterns are controlled as 40$\ast$40 matrices of black and white squares by computer. And the ratio of white and black squares is 0.11 in the patterns. The changing speckle patterns are projected onto the object. The distance between the projector and object is 30 cm. The transmitted light through the letters is reflected by the rough white wall. The distance between the object and the wall is 15 cm. The single-pixel detector, which is 12 cm away from the wall, collects the reflected light. A light tight baffle is inserted between the object and the single-pixel detector to avoid the detector receiving the signal from the object directly. The result is shown in Fig. \ref{NLOSXJTU}(a), where the letters are imaged clearly. To study the ability of this camera against scattering, a rotating ground glass (5 cm behind the object) is placed between the object and the detector. The rotating speed of the ground glass is 1,200 cycles/min. The result is shown in Fig. \ref{NLOSXJTU}(b). The SNR of the image in Fig. \ref{NLOSXJTU}(a) and Fig. \ref{NLOSXJTU}(b) are 17.34 \emph{dB} and 17.51 \emph{dB}, respectively. They indicate that imaging around corners with the single-pixel detector can image clearly whether there is the scattering media or not. Especially, we use white paper as the distraction. Two pieces of paper are placed at the different locations behind the object. One is 120 mm behind the object and the other is 10 mm in front of the detector. The result is shown in Fig. \ref{NLOSXJTU}(c) and the SNR is 18.58 \emph{dB}.

\begin{figure}[htb]
    \centering
    \includegraphics[width=80mm]{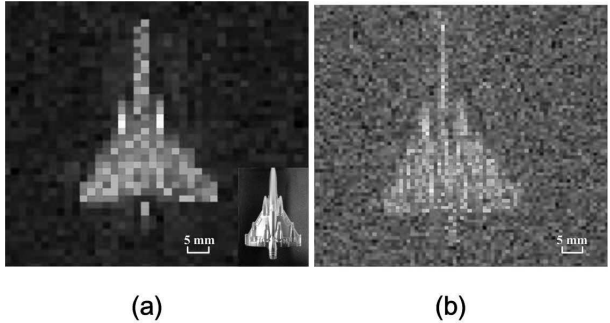}
    \caption{ The image and object in the experiment of imaging around corners with the single-pixel detector. The inset picture in (a) is the diffuse object (a common toy for children). (a) and (b) show the result of image. The number of patterns which is used to image in two pictures are both 50,000. The resolution ratio of (b) is 4 times better than (a). }
    \label{toy}
\end{figure}

Figure \ref{NLOS1}(c) shows the scheme of imaging the diffuse object by the single-pixel camera with imaging around corners. A three-dimensional object with rough surface is used as the imaging target. The inset in Fig. \ref{toy}(a) is this diffuse object (a common toy plane for children). The size of object is 55 mm$\ast$32 mm. Speckle patterns are irradiated on this diffuse object. The intensity of light reflected from the object and the wall is very weak. In addition, the environment light is significantly strong. Under the circumstances, the detector will receive more light from environment than from the object. Traditional camera generally gets a picture with the misty light. The light becomes disordered and the image is blurred because of the reflection on the diffuse object and the wall. However, the single-pixel camera with imaging around corners can image the object under the same condition. The result of imaging by the camera with imaging around corners is shown in Fig. \ref{toy}(a), where the toy is imaged clearly. The SNR of the image in Fig. \ref{toy}(a) is 23 \emph{dB}. The diffuse object is more difficult to be imaged than the simple transmission object, therefore the number of speckle patterns is increased to 50,000 to get a clear image. In addition, the resolution can be controlled by the computer. When the size of speckle patterns becomes smaller, the resolution of the image becomes higher. Fig. \ref{toy}(b) is the result when the resolution increases to 4 times better than (a). All the details of the object in Fig. \ref{toy}(b) become more obvious. The SNR of the image becomes 20 \emph{dB}. All results show that the single-pixel camera with imaging around corners can image the diffuse object when the camera cannot look at object directly. 

\section{Theory and Discussion}

\begin{figure}[htb]
    \centering
    \includegraphics[width=70 mm]{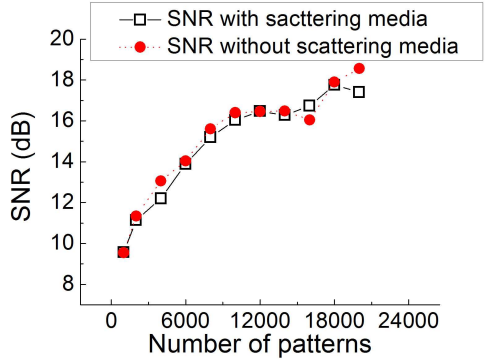}
    \caption{ The SNR of images. The SNR of the images of the transparent object in CGI is measured in two cases. One case, which is represented as square, is that the scattering media exists. The other case, which is represented as circle, is that the scattering media does not exist. The tendency of SNR with the increasing number of speckle patterns is shown. }
    \label{SNR}
\end{figure}

The image of the object in CGI is obtained via calculating the second-order correlation,

\begin{equation}\label{g2}
g^{(2)}(x,y)=\frac{\left \langle I_{com}(x,y)I_{det}\right \rangle}{\left \langle I_{com}(x,y)\right \rangle \left \langle I_{det} \right \rangle},
\end{equation}
where $\langle\rangle$ means ensemble average. $I_{com}(x,y)$ is the intensity distribution of speckle patterns which are controlled by computer. $I_{det}$ is the intensity detected by the single-pixel detector.

\begin{equation}\label{g2ob}
I_{det}=\int\!\!\!\int I_{obj}(x',y')T(x',y')\mathrm{d}x'\mathrm{d}y',
\end{equation}
where $I_{obj}(x',y')$  is the intensity of light in the object plane. $T(x',y')$ is the transmission function of the object. Equation (\ref{g2}) can be simplified as \cite{4}
\begin{equation}\label{g21}
g^{(2)}(x,y)=1+\frac{\left \langle \Delta I_{com}(x,y)\Delta I_{det}\right \rangle}{\left \langle I_{com}(x,y)\right \rangle \left \langle I_{det} \right \rangle},
\end{equation}
where $\Delta I_{com}(x,y)$ and $\Delta I_{det}$ are the intensity fluctuations to their ensemble averages, respectively. It shows that $g^{(2)}(x,y)$  depends on the second-order intensity fluctuation correlation $\left \langle \Delta I_{com}(x,y)\Delta I_{det}\right \rangle$. Light first propagates from the source to one point on object, and then through this point arrives in the detector plane. When the intensity fluctuation of the detector signal contains the intensity fluctuations of light through this point on the object, the value of $g^{(2)}(x,y)$ on the same location of that point in the picture would be greater than 1. At last, the object would be imaged on the picture. When a ground glass is inserted between the object and the detector, light is scattered in all directions. If the scattering is strong enough, the intensity of the transmitted light through all points on the object would be close to homogeneous distribution in space. The single-pixel detector can only obtain parts of light which is fully scattered. Furthermore, the light intensity is relatively weak. Nevertheless, this part of light still contains all information of object, which means that the single-pixel detector can get the information of the intensity fluctuations of light from every points on the object. In addition, the signal of detector $I_{det}$ also contains the intensity of light from the ground glass. Because the intensity from the glass and that from computer are independent, there is no correlation between them. The effect from the ground glass will disappear when $ \left \langle \Delta I_{com}(x,y)\Delta I_{det}\right \rangle $ is calculated \cite{38}. The imaging of the object can still be recovered successfully. However, the visibility of the image will become lower since the background $ \left \langle I_{com}(x,y)\right \rangle \left \langle I_{det} \right \rangle $ relatively increases. According to the above equations and discussions, CGI can still image letters of XJTU through the scattering media. Figure \ref{XJTU}(b) shows the result with scattering media is similar as the image in CGI without scattering media.
Furthermore, the SNR of the image of the transparent object in CGI is studied. Figure \ref{SNR} shows the SNR of the image changes with the number of collected patterns in both cases. The difference of two cases is whether the scattering media exists. Under two different conditions, the curves have a good coincide. It illustrates that the scattering media behind object has little effect to the quality of the image.

When the scattering media becomes a reflection type, imaging around corners with the single-pixel detector can be achieved. The transmitted light through the object reaches the white wall which replaces the former ground glass. Light with the information of the object will be scattered into every direction in the room, so the information of the object can be obtained in many places. $I_{det}$ will contain the intensity of light reflected from the object, the white wall and optical noise. However, there is no correlation between the intensity of light controlled by computer and all intensities of light except reflected from the object. It will be the same calculation as the one of CGI with scattering media if the transmission function of the scattering media is replaced by the reflection function of the white wall. The image of the object will be shown in the area where $g^{(2)}(x,y)$ is greater than 1. Imaging through scattering layers and around corners is also studied by some researchers \cite{auto,giuliano}. They applied the memory-effect for speckle correlations \cite{auto} and the Fourier-domain shower-curtain effect \cite{giuliano} to image the object. The difference between the strategies is that our image system is a kind of ``direct imaging'' which is based on simple product of intensities of a bucket detector and preset speckles patterns. It does not need a complex computation such as phase retrieval algorithms. The image only needs that the correlation is calculated. Moreover, the detector in our experiment is a single-pixel detector which is simpler than the high-megapixel camera.

When a diffuse object instead of the transmission object is imaged, the intensity would multiply by the reflection function instead of the transmission function of the object in Eq.(\ref{g2ob}). Light with the information of the object will be scattered again. The information of the object can be obtained in many places even though the intensity of light is still extremely weak. Under this condition, there is still correlation between the intensity of light controlled by computer and the intensity of light from the object. In the end, the image of object will appear as if the sight of the single-pixel camera with imaging around corners can bypass the barrier.

\section{Conlusions}

In conclusion, we designed and accomplished the imaging around corners with the single-pixel detector by a series of experiments about CGI with a single-pixel detector. This imaging around corners with the single-pixel detector obtains parts of light reflected from the wall and cannot ``see'' the diffuse object directly. However, it still can obtain the image of the object. We analyzed the feature of the camera without image optimisation, so this camera still has  great potential in enhancing image quality and speed. With optimized algorithms to make CGI faster and the experimental equipments connected by a wireless system, the single-pixel camera with imaging around corners can be a powerful equipment in imaging around corners and obstacles in real world.

\section*{Acknowledgement}
This work is supported by the National Basic Research Program of China (973 Program) under Grant  No.2015CB654602 and the 111 Project of China under Grant No. B14040.

\end{document}